\documentclass[letterpaper]{article} 
\usepackage{aaai25}  
\usepackage{times}  
\usepackage{helvet}  
\usepackage{courier}  
\usepackage[hyphens]{url}  
\usepackage{graphicx} 
\usepackage{natbib}  
\usepackage{caption} 
\usepackage{algorithm}
\usepackage{algorithmic}

\usepackage{newfloat}
\usepackage{listings}
\DeclareCaptionStyle{ruled}{labelfont=normalfont,labelsep=colon,strut=off} 
\floatstyle{ruled}
\newfloat{listing}{tb}{lst}{}
\floatname{listing}{Listing}
\usepackage[utf8]{inputenc} 
\usepackage[T1]{fontenc}    
\usepackage{url}            
\usepackage{booktabs}       
\usepackage{amsfonts}       
\usepackage{nicefrac}       
\usepackage{microtype}      
\usepackage{xcolor}         

\usepackage[utf8]{inputenc} 
\usepackage[T1]{fontenc}    
\usepackage{url}            
\usepackage{booktabs}       
\usepackage{amsfonts}       
\usepackage{nicefrac}       
\usepackage{microtype}      
\usepackage{graphicx}
\usepackage{subfigure}
\usepackage{booktabs} 
\usepackage{amsmath, amsthm, amsfonts}

\usepackage{mathtools}
\usepackage{algorithmic, algorithm}
\usepackage[belowskip=0pt, aboveskip=0pt]{caption}
\usepackage{multirow}

\usepackage{verbatim}
\usepackage{xcolor}
\usepackage[font=footnotesize,labelfont=bf]{caption}

\graphicspath{ {./Figures/} }

\title{The Influence of Initial Connectivity on Biologically Plausible Learning}

\author{
    Weixuan Liu\textsuperscript{\rm 1,*},
    Xinyue Zhang\textsuperscript{\rm 2,*},  
    Yuhan Helena Liu\textsuperscript{\rm 3,+}
}
\affiliations{
    \textsuperscript{\rm 1}University of Washington, Seattle, WA, USA\\

    \textsuperscript{\rm 2}Columbia University, New York, NY, USA\\
    \textsuperscript{\rm 3}Princeton University, Princeton, NJ, USA\\
    \textsuperscript{\rm *}Joint first authors, \textsuperscript{\rm +}Correspondence: hl7582@princeton.edu
}

\begin{document}

\maketitle

\begin{abstract}
Understanding how the brain learns can be advanced by investigating biologically plausible learning rules --- those that obey known biological constraints, such as locality, to serve as valid brain learning models. Yet, many studies overlook the role of architecture and initial synaptic connectivity in such models. Building on insights from deep learning, where initialization profoundly affects learning dynamics, we ask a key but underexplored neuroscience question: how does initial synaptic connectivity shape learning in neural circuits? To investigate this, we train recurrent neural networks (RNNs), which are widely used for brain modeling, with biologically plausible learning rules. Our findings reveal that initial weight magnitude significantly influences the learning performance of such rules, mirroring effects previously observed in training with backpropagation through time (BPTT). By examining the maximum Lyapunov exponent before and after training, we uncovered the greater demands that certain initialization schemes place on training to achieve desired information propagation properties. Consequently, we extended the recently proposed gradient flossing method, which regularizes the Lyapunov exponents, to biologically plausible learning and observed an improvement in learning performance. To our knowledge, we are the first to examine the impact of initialization on biologically plausible learning rules for RNNs and to subsequently propose a biologically plausible remedy. Such an investigation can lead to neuroscientific predictions about the influence of initial connectivity on learning dynamics and performance, as well as guide neuromorphic design. 

\end{abstract}

\section{Introduction}

A central question in computational neuroscience is how initial connectivity influences the dynamics of learning. While the magnitude of initial weights is known to influence these dynamics in backpropagation-based gradient descent learning~\cite{flesch2021rich,chizat2019lazy,schuessler2020interplay,braun2022exact,woodworth2020kernel,paccolat2021geometric,schuessler2023aligned}, the neural implementation challenges of backpropagation~\cite{lillicrap2020backpropagation,richards2019deep,lillicrap2019backpropagation,hinton2022forward} raise important questions about its validity as a neural learning model and how such influences extend to biologically plausible learning. This inquiry is especially relevant for recurrent neural networks (RNNs), which are widely employed in modeling neural circuits~\cite{yang2020artificial,molano2022neurogym,vyas2020computation}.

Understanding how the brain learns can be advanced by investigating biologically plausible (bio-plausible) learning rules, which aim to capture the interactions among neural components that enable learning while adhering to known biological constraints, such as locality, where all mathematical terms involved in weight updates can be mapped onto known biological signals that are physically present at the synapse~\cite{marschall2019evaluating}. These rules have been a focus of recent computational neuroscience efforts to model learning~\cite{lillicrap2020backpropagation,richards2019deep}.

In light of this, we ask: How does the initialization of weights, particularly their magnitude, affect the performance of biologically plausible learning in RNNs? We evaluate performance primarily through learning curves, measured by the reduction in loss over training. Our focus is on biologically plausible learning rules that approximate gradients by truncating non-biological terms, specifically the two equivalent rules of e-prop and random feedback local online (RFLO) learning, which have shown efficacy and versatility in solving complex tasks~\cite{murray2019local,bellec2020solution}. 

Our contributions are as follows: \textbf{(1)} We demonstrate that, much like in BPTT, the initial weight magnitude in e-prop significantly affects learning performance (Figure~\ref{fig:main_learning_curve}). \textbf{(2)} To explain this result, we identified that the maximum Lyapunov exponent --- crucial for the stability of information propagation --- undergoes the most significant changes with small initial weight magnitudes, suggesting greater demands are placed on training (Figure~\ref{fig:maxLE_gain}). \textbf{(3)} Consequently, we extended the recently proposed gradient flossing method~\cite{engelken2024gradient} --- designed to stability training by regularizing Lyapunov exponents --- to the context of biologically plausible learning; this improved the performance significantly (Figure~\ref{fig:grad_floss}), particularly when the initial magnitude was suboptimal, which might occur due to pathological conditions. 

\section{Results}

\subsection{Network and training setup}

\begin{figure*}[t]
    \centering
    \includegraphics[width=0.8\textwidth]{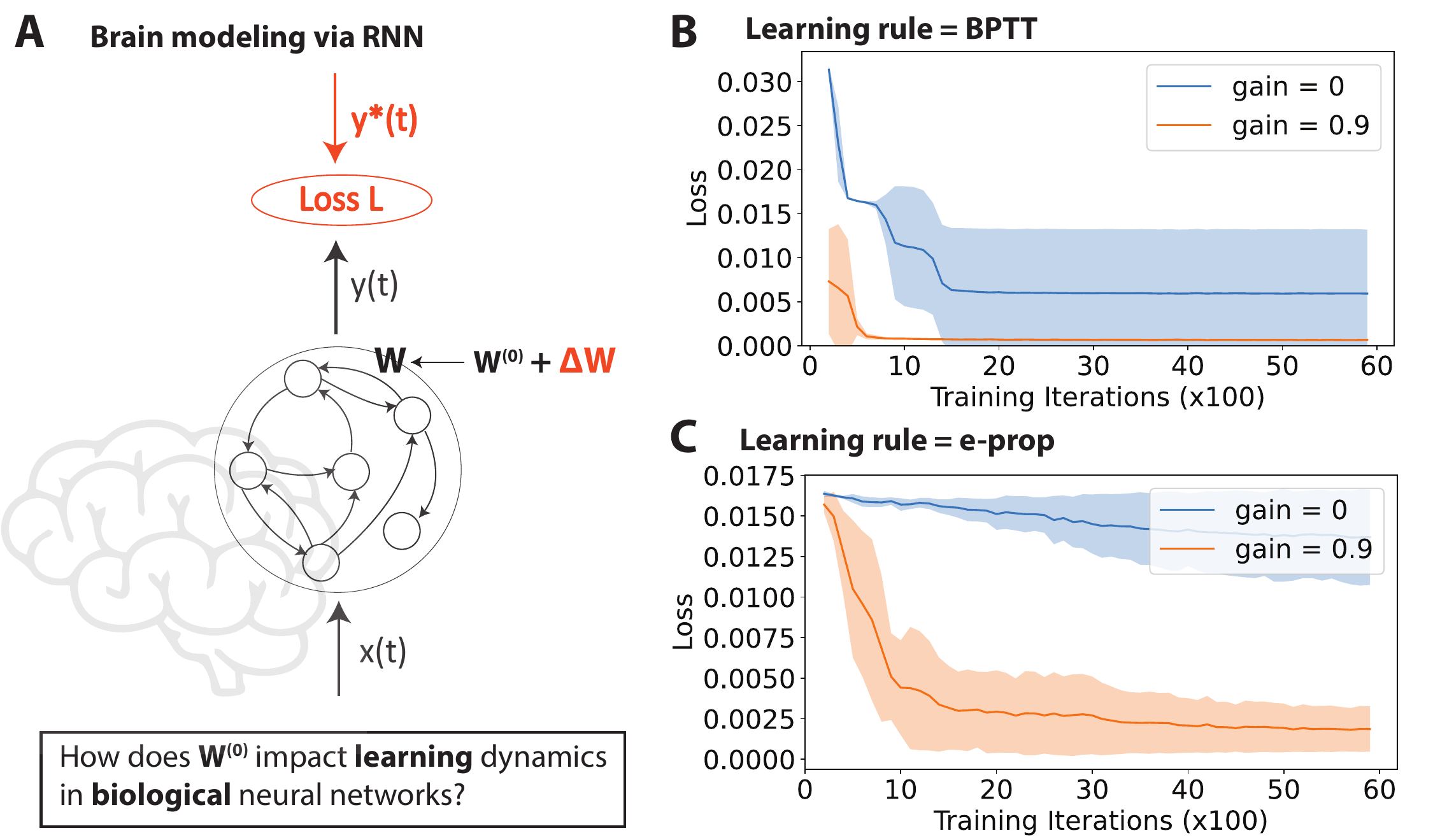}
    \caption{ \textbf{Influence of weight initialization on biologically plausible learning}. A) RNN Setup for Brain Learning Models. B) Loss curves throughout training for different initial gains using the standard backpropagation through time (BPTT) algorithm. Here, gain reflects the initial weight magnitude: recurrent weights are initialized as $W^{(0)}_{h,ij} \sim \mathcal{N}(0, gain^2/N)$. C) Loss curve throughout training for different initial gains using e-prop, a biologically plausible learning rule for RNNs. Note: the plots in B) and C) begin after 200 training iterations to provide a more focused view of the results. This figure illustrates the Romo task but similar trends are observed for the 2AF and DMS tasks (Figure~\ref{fig:more_tasks}). Learning curves for more intermediate gain values are examined in Appendix Figure~\ref{fig:more_gains}. Solid lines/shaded regions: mean/standard deviation of loss curves across independent runs with different seeds. 
    }
    \label{fig:main_learning_curve}
\end{figure*}

\begin{figure*}[t]
    \centering
    \includegraphics[width=0.8\textwidth]{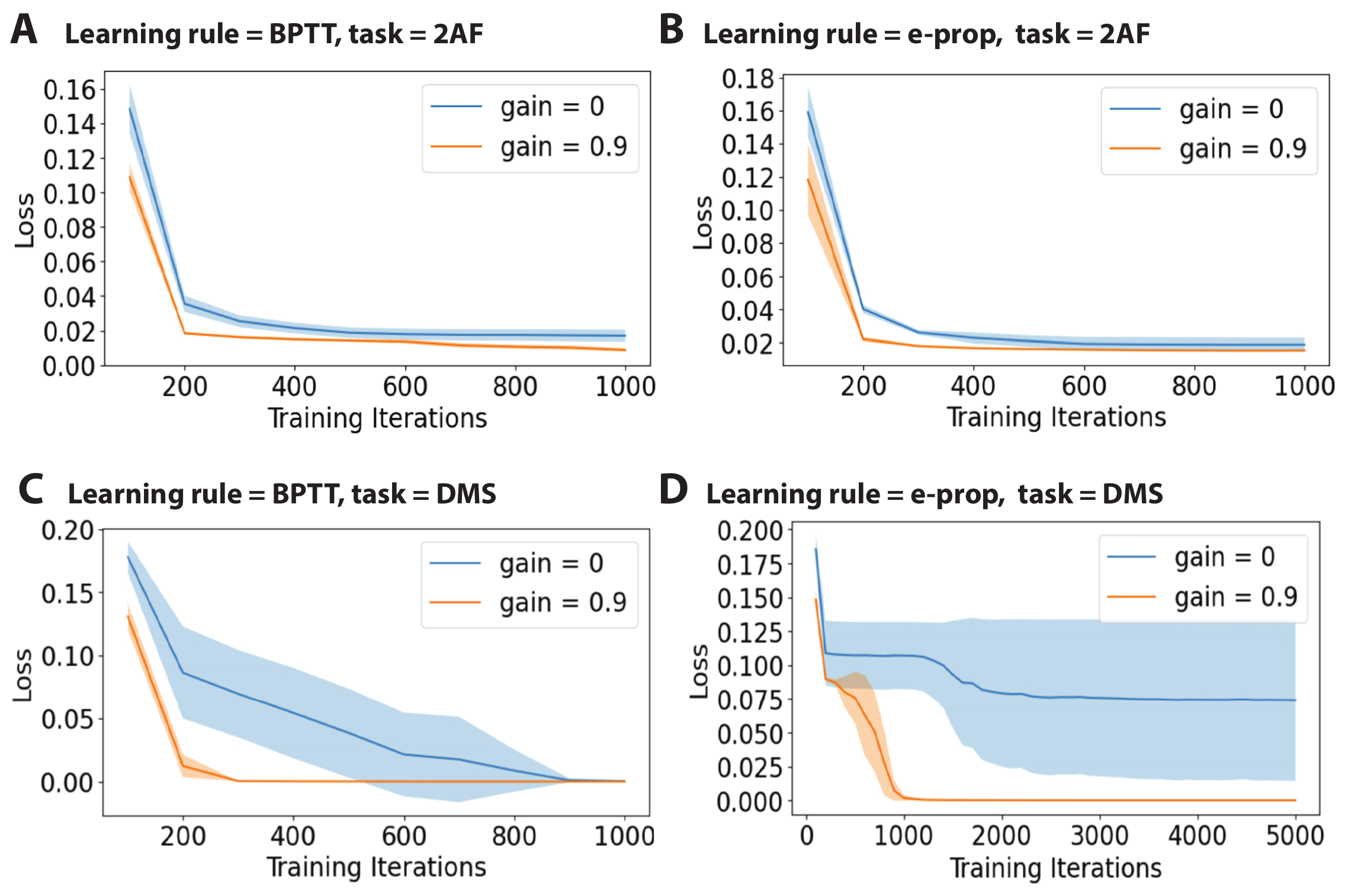}
    \caption{ \textbf{Similar trends as Figure~\ref{fig:main_learning_curve} observed for the 2AF and DMS tasks}. A) Learning curve for the 2AF task across different initial gains using backpropagation through time (BPTT). B) Similar to A) but for e-prop. C) Similar to A) but for the DMS task. D) Similar to B) but for the DMS task. Plotting convention follows that of Figure~\ref{fig:main_learning_curve}. 
    }
    \label{fig:more_tasks}
\end{figure*}

We examine recurrent neural networks (RNNs) because they are commonly adopted for modeling neural circuits \cite{barak2017recurrent, song2016training}. Our RNN model (Figure~\ref{fig:main_learning_curve}A) comprises $N_{in}$ input nodes, $N$ hidden nodes, and $N_{out}$ output nodes. The hidden state at time $t$, denoted as $h_t \in \mathbb{R}^N$, is updated according to the following equation: 
\begin{equation}
h_{t+1} = \alpha h_t + (1-\alpha) (W_h f(h_t) + W_x x_t),
\end{equation}
where the leak factor $\alpha = 1 - \frac{dt}{\tau} \in \mathbb{R}$ is determined by the simulation time step $dt$ and the membrane time constant $\tau$. The function $f(\cdot): \mathbb{R}^N \rightarrow \mathbb{R}^N$ is the $ReLU$ activation function; $W_h \in \mathbb{R}^{N \times N}$ and $W_x \in \mathbb{R}^{N \times N_{in}}$ represent the recurrent and input weight matrices, respectively; and $x_t \in \mathbb{R}^{N_{in}}$ is the input at time $t$. The output, $\hat{y}_{t} \in \mathbb{R}^{N{out}}$, is derived as a linear combination of the hidden state activation, $f(h_t)$, using the readout weights $w \in \mathbb{R}^{N_{out} \times N}$.

The goal is to minimize the scalar loss $L \in \mathbb{R}$. For loss minimization, we explored several learning rules, including BPTT, which calculates the exact gradient, $\nabla_W L(W_h) \in \mathbb{R}^{N \times (N_{in} + N + N_{out})}$, as well as biologically plausible learning rules that utilize approximate gradients, $\widehat{\nabla_W} L(W) \in \mathbb{R}^{N \times (N_{in} + N + N_{out})}$:
\begin{equation}
\Delta W = -\eta \nabla_{W} L(W),
\end{equation}
\begin{equation}
\widehat{\Delta W} = -\eta \widehat{\nabla_W} L(W),
\end{equation}
where $W = [W_h \quad W_x \quad w^T] \in \mathbb{R}^{N \times (N_{in} + N + N_{out})}$ represents all the trainable parameters, and $\eta $ is the learning rate.

In the realm of biologically plausible learning rules for RNNs, we focused primarily on e-prop \cite{bellec2020solution} and RFLO \cite{murray2019local}, which rely on gradient truncation. Since both are equivalent in our setting, we present only the results for e-prop. A significant challenge with the neural implementation of BPTT arises from its weight updates, which require precise gradients of the loss with respect to the weights. This process demands that every synapse receive activity signals from the entire recurrent network \cite{marschall2020unified}, a mechanism that raises serious questions about its validity for modeling neural circuit learning. In contrast, e-prop and RFLO truncate this exact gradient, ensuring that the remaining terms can be associated with known biological processes; specifically, the weight update depends on the pre- and postsynaptic activities along with a third factor that guides the weight update. Although other biologically plausible learning rules exist, we concentrated on e-prop and RFLO due to their versatility and being the focus of recent studies examining RNN learning rules~\cite{liu2022beyond,portes2022distinguishing}. For example, rules like equilibrium propagation depend on the equilibrium condition \cite{scellier2017equilibrium,meulemans2022least}.

We simulated different neuroscience tasks. In the main text, we displayed results for the Romo task~\cite{romo1999neuronal}, following the implementation in~\cite{schuessler2020interplay}, but also showed the trend applies to other tasks --- including perceptual decision-making (2AF) and the delayed-match-to-sample (DMS) tasks --- implemented using Neurogym \cite{molano2022neurogym} (Figure~\ref{fig:more_tasks}). Training details as well as additional explanations on gradient flossing and the learning rules can be found in the Appendix. 

\subsection{Simulation results}

\begin{figure*}[h!]
    \centering
    \includegraphics[width=0.8\textwidth]{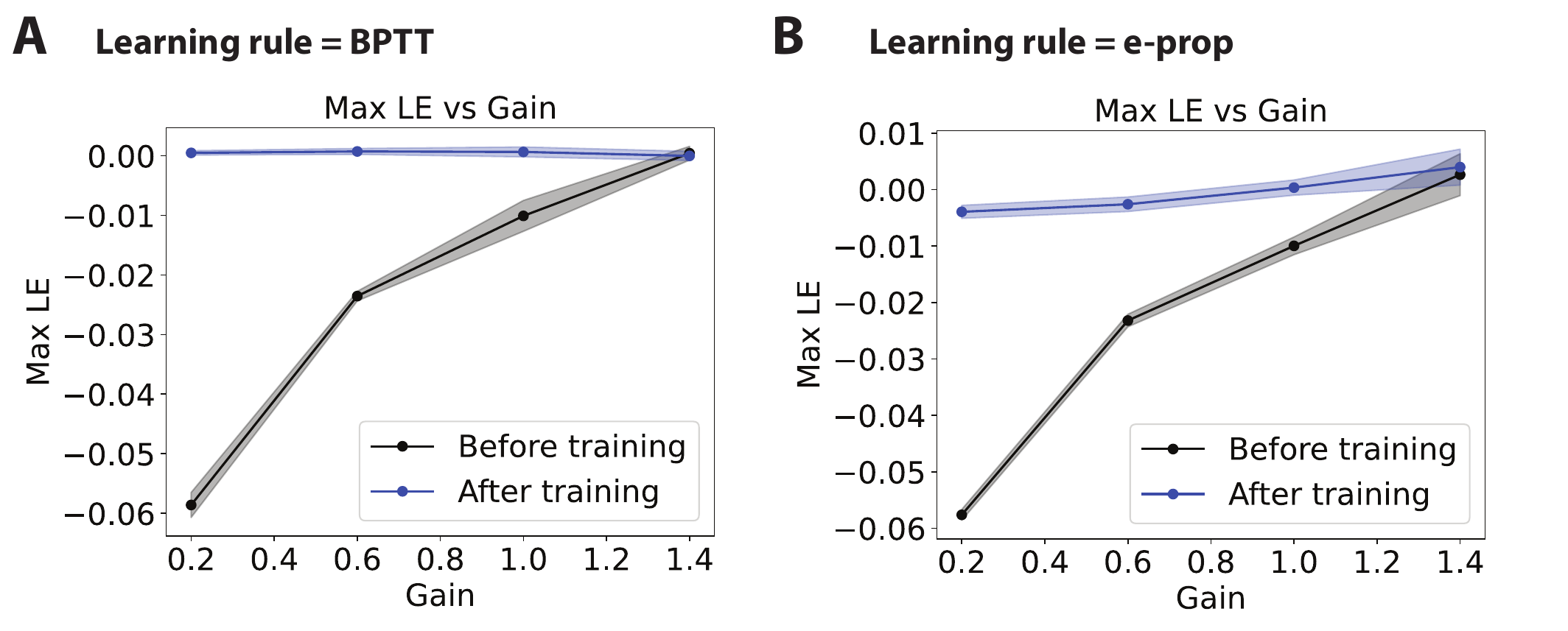}
    \caption{ The maximum Lyapunov exponent (Max LE)  computed before and after training across various weight initialization gains for training via A) BPTT and B) e-prop. Gain is defined similarly as in Figure~\ref{fig:main_learning_curve}. Certain initial weight magnitudes result in more significant changes in the Max LE. Solid lines/shaded regions: mean/standard deviation of Max LE across independent runs with different seeds.
    }
    \label{fig:maxLE_gain}
\end{figure*}

\begin{figure*}[h!]
    \centering
    \includegraphics[width=0.8\textwidth]{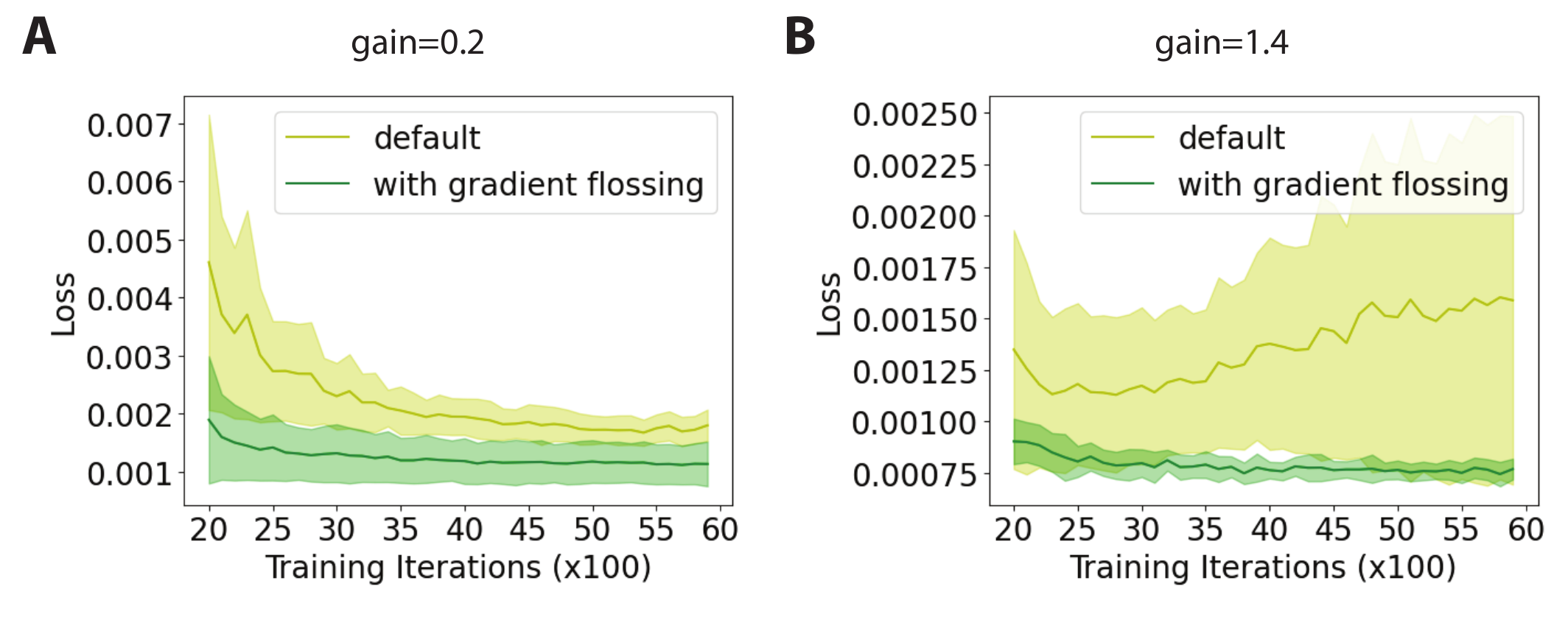}
    \caption{ Initialization via pretraining with gradient flossing improves e-prop learning performance, particularly for suboptimal initialization gains. Note: The plot begins after 2000 iterations to provide a more focused view of the results.
    }
    \label{fig:grad_floss}
\end{figure*}

We examined the effects of different initial weight magnitudes, which have been shown to significantly influence the learning trajectory and final solution in BPTT~\cite{schuessler2020interplay}. Figure~\ref{fig:main_learning_curve} demonstrates that the performance gap, as indicated by the learning curve, is substantial across different initialization magnitudes for both BPTT and e-prop. Additional intermediate magnitudes are explored in Appendix Figure~\ref{fig:more_gains}, where notable gap is observed for certain initial weight magnitudes. Similar trends are evident when the experiments are repeated across other tasks, specifically the 2AF and DMS tasks implemented using Neurogym (Figure~\ref{fig:more_tasks}). These results underscore the critical role of weight initialization in biologically plausible learning.

Next, we investigate why initialization has such a profound effect on learning performance in biologically plausible learning. We turn to Lyapunov exponents, which can reflect the ability of RNNs to propagate information~\cite{vogt2022lyapunov}. Lyapunov exponents help in studying the dynamical properties of RNNs, as they measure the system's sensitivity to initial conditions and quantify the rates of divergence or convergence of trajectories in the system's state space. We computed the Lyapunov exponent using the method described in~\cite{vogt2022lyapunov} for networks before and after training. The analysis was done for the Romo task but similar trends were observed for other tasks as well. As expected, the trained networks exhibit a maximum Lyapunov exponent around $0$, so that the signals neither explode nor vanish. However, before training, networks initialized with smaller weight gains had Lyapunov exponents further from $0$, indicating that more changes are required via training, thus making the process more challenging for such initializations (Figure~\ref{fig:maxLE_gain}).

To address this, we applied the recently proposed gradient flossing method~\cite{engelken2024gradient}, which adjusts Lyapunov exponents closer to $0$ and has been shown to improve BPTT training performance. We adapted this method for biologically plausible learning by pretraining the network with 100 iterations using the "flossing loss" while ensuring the weight updates use local information only (see Appendix). Our results show that this approach of gradient flossing also enhances performance in this context of biologically plausible learning, particularly when the initial weight gain is suboptimal (Figure~\ref{fig:grad_floss}), which might happen due to pathology.

\section{Discussion}

This study highlights the role of initial weight magnitude in shaping the learning dynamics of biologically plausible rules, predicting its importance in neural circuit learning. While the influence of initial connectivity on learning has been extensively explored in the realm of backpropagation-based learning, our work is novel because it extends this inquiry to biologically plausible settings. Our findings demonstrate that, similar to backpropagation through time (BPTT), the choice of initial weight magnitude in e-prop --- a biologically plausible learning rule --- has a profound impact on learning performance. Notably, we observed that smaller initial gains can paradoxically hinder learning. This result is explained by our analysis of the Lyapunov exponent, which is crucial for the stability and information propagation within the network. We found that smaller initial gains resulted in larger deviations of the Lyapunov exponent from zero before training, indicating a greater challenge in achieving the balanced dynamical properties necessary for effective learning. To address this challenge, we brought the gradient flossing method into the biologically plausible learning framework, leading to performance improvement for suboptimal initial weight magnitudes. Overall, these findings provide insights into how variations in initial connectivity may influence learning in neural circuits, offering predictions that can guide future experimental work. Additionally, these findings have practical implications for the design of neuromorphic chips, where optimizing initial weight configurations could enhance the efficiency and effectiveness of energy-efficient biologically plausible learning algorithms.

Extending our approach to explore the interaction between initialization and biologically plausible learning rules across a broader range of learning rules, architectures, and tasks is an important direction for future research. In this study, we focused on existing biologically plausible RNN learning rules~\cite{murray2019local,bellec2020solution,liu2021cell}, chosen for their demonstrated efficacy in task learning, versatility in settings (e.g., avoiding the equilibrium assumption~\cite{scellier2017equilibrium,meulemans2022least}), and prominence in recent computational neuroscience studies~\cite{liu2022beyond,portes2022distinguishing}. An important future direction would involve exploring a wider range of learning rules as well as paradigms, including reinforcement learning~\cite{sutton2018reinforcement}, beyond the supervised learning setup currently examined. Moreover, while we examined the magnitude of initial connectivity due to its known influence on BPTT-based learning dynamics~\cite{schuessler2020interplay}, other attributes of initialization may also play critical roles~\cite{liu2023connectivity}. Future work could investigate these factors along with other aspects, such as the interaction between rich and lazy learning regimes and their impact on generalization~\cite{chizat2019lazy,jacot2018neural}. 

\subsubsection{Acknowledgments.}
This research was initiated and supported through the WAMM Program at the University of Washington. Y.H.L. is funded via NSERC PGS-D, FRQNT B2X and B3X, and the Pearson Fellowship. 

\bibliography{ref_main}

\begin{thebibliography}{31}
\providecommand{\natexlab}[1]{#1}

\bibitem[{Barak(2017)}]{barak2017recurrent}
Barak, O. 2017.
\newblock Recurrent neural networks as versatile tools of neuroscience
  research.
\newblock \emph{Current opinion in neurobiology}, 46: 1--6.

\bibitem[{Bellec et~al.(2020)Bellec, Scherr, Subramoney, Hajek, Salaj,
  Legenstein, and Maass}]{bellec2020solution}
Bellec, G.; Scherr, F.; Subramoney, A.; Hajek, E.; Salaj, D.; Legenstein, R.;
  and Maass, W. 2020.
\newblock A solution to the learning dilemma for recurrent networks of spiking
  neurons.
\newblock \emph{Nature communications}, 11(1): 3625.

\bibitem[{Braun et~al.(2022)Braun, Domin{\'e}, Fitzgerald, and
  Saxe}]{braun2022exact}
Braun, L.; Domin{\'e}, C.; Fitzgerald, J.; and Saxe, A. 2022.
\newblock Exact learning dynamics of deep linear networks with prior knowledge.
\newblock \emph{Advances in Neural Information Processing Systems}, 35:
  6615--6629.

\bibitem[{Chizat, Oyallon, and Bach(2019)}]{chizat2019lazy}
Chizat, L.; Oyallon, E.; and Bach, F. 2019.
\newblock On lazy training in differentiable programming.
\newblock \emph{Advances in neural information processing systems}, 32.

\bibitem[{Engelken(2024)}]{engelken2024gradient}
Engelken, R. 2024.
\newblock Gradient flossing: Improving gradient descent through dynamic control
  of jacobians.
\newblock \emph{Advances in Neural Information Processing Systems}, 36.

\bibitem[{Flesch et~al.(2021)Flesch, Juechems, Dumbalska, Saxe, and
  Summerfield}]{flesch2021rich}
Flesch, T.; Juechems, K.; Dumbalska, T.; Saxe, A.; and Summerfield, C. 2021.
\newblock Rich and lazy learning of task representations in brains and neural
  networks.
\newblock \emph{BioRxiv}, 2021--04.

\bibitem[{Hinton(2022)}]{hinton2022forward}
Hinton, G. 2022.
\newblock The forward-forward algorithm: Some preliminary investigations.
\newblock \emph{arXiv preprint arXiv:2212.13345}.

\bibitem[{Jacot, Gabriel, and Hongler(2018)}]{jacot2018neural}
Jacot, A.; Gabriel, F.; and Hongler, C. 2018.
\newblock Neural tangent kernel: Convergence and generalization in neural
  networks.
\newblock \emph{Advances in neural information processing systems}, 31.

\bibitem[{Lillicrap and Santoro(2019)}]{lillicrap2019backpropagation}
Lillicrap, T.~P.; and Santoro, A. 2019.
\newblock Backpropagation through time and the brain.
\newblock \emph{Current opinion in neurobiology}, 55: 82--89.

\bibitem[{Lillicrap et~al.(2020)Lillicrap, Santoro, Marris, Akerman, and
  Hinton}]{lillicrap2020backpropagation}
Lillicrap, T.~P.; Santoro, A.; Marris, L.; Akerman, C.~J.; and Hinton, G. 2020.
\newblock Backpropagation and the brain.
\newblock \emph{Nature Reviews Neuroscience}, 21(6): 335--346.

\bibitem[{Liu et~al.(2023)Liu, Baratin, Cornford, Mihalas, Shea-Brown, and
  Lajoie}]{liu2023connectivity}
Liu, Y.~H.; Baratin, A.; Cornford, J.; Mihalas, S.; Shea-Brown, E.; and Lajoie,
  G. 2023.
\newblock How connectivity structure shapes rich and lazy learning in neural
  circuits.
\newblock \emph{ArXiv}.

\bibitem[{Liu et~al.(2022)Liu, Ghosh, Richards, Shea-Brown, and
  Lajoie}]{liu2022beyond}
Liu, Y.~H.; Ghosh, A.; Richards, B.; Shea-Brown, E.; and Lajoie, G. 2022.
\newblock Beyond accuracy: generalization properties of bio-plausible temporal
  credit assignment rules.
\newblock \emph{Advances in Neural Information Processing Systems}, 35:
  23077--23097.

\bibitem[{Liu et~al.(2021)Liu, Smith, Mihalas, Shea-Brown, and
  S{\"u}mb{\"u}l}]{liu2021cell}
Liu, Y.~H.; Smith, S.; Mihalas, S.; Shea-Brown, E.; and S{\"u}mb{\"u}l, U.
  2021.
\newblock Cell-type--specific neuromodulation guides synaptic credit assignment
  in a spiking neural network.
\newblock \emph{Proceedings of the National Academy of Sciences}, 118(51):
  e2111821118.

\bibitem[{Marschall, Cho, and Savin(2019)}]{marschall2019evaluating}
Marschall, O.; Cho, K.; and Savin, C. 2019.
\newblock Evaluating biological plausibility of learning algorithms the lazy
  way.
\newblock In \emph{Real Neurons $\{$$\backslash$\&$\}$ Hidden Units: Future
  directions at the intersection of neuroscience and artificial intelligence@
  NeurIPS 2019}.

\bibitem[{Marschall, Cho, and Savin(2020)}]{marschall2020unified}
Marschall, O.; Cho, K.; and Savin, C. 2020.
\newblock A unified framework of online learning algorithms for training
  recurrent neural networks.
\newblock \emph{The Journal of Machine Learning Research}, 21(1): 5320--5353.

\bibitem[{Meulemans et~al.(2022)Meulemans, Zucchet, Kobayashi, Von~Oswald, and
  Sacramento}]{meulemans2022least}
Meulemans, A.; Zucchet, N.; Kobayashi, S.; Von~Oswald, J.; and Sacramento, J.
  2022.
\newblock The least-control principle for local learning at equilibrium.
\newblock \emph{Advances in Neural Information Processing Systems}, 35:
  33603--33617.

\bibitem[{Molano-Mazon et~al.(2022)Molano-Mazon, Barbosa, Pastor-Ciurana,
  Fradera, Zhang, Forest, del Pozo~Lerida, Ji-An, Cueva, de~la Rocha
  et~al.}]{molano2022neurogym}
Molano-Mazon, M.; Barbosa, J.; Pastor-Ciurana, J.; Fradera, M.; Zhang, R.-Y.;
  Forest, J.; del Pozo~Lerida, J.; Ji-An, L.; Cueva, C.~J.; de~la Rocha, J.;
  et~al. 2022.
\newblock NeuroGym: An open resource for developing and sharing neuroscience
  tasks.

\bibitem[{Murray(2019)}]{murray2019local}
Murray, J.~M. 2019.
\newblock Local online learning in recurrent networks with random feedback.
\newblock \emph{Elife}, 8: e43299.

\bibitem[{Paccolat et~al.(2021)Paccolat, Petrini, Geiger, Tyloo, and
  Wyart}]{paccolat2021geometric}
Paccolat, J.; Petrini, L.; Geiger, M.; Tyloo, K.; and Wyart, M. 2021.
\newblock Geometric compression of invariant manifolds in neural networks.
\newblock \emph{Journal of Statistical Mechanics: Theory and Experiment},
  2021(4): 044001.

\bibitem[{Portes, Schmid, and Murray(2022)}]{portes2022distinguishing}
Portes, J.; Schmid, C.; and Murray, J.~M. 2022.
\newblock Distinguishing learning rules with brain machine interfaces.
\newblock \emph{Advances in neural information processing systems}, 35:
  25937--25950.

\bibitem[{Richards et~al.(2019)Richards, Lillicrap, Beaudoin, Bengio, Bogacz,
  Christensen, Clopath, Costa, de~Berker, Ganguli et~al.}]{richards2019deep}
Richards, B.~A.; Lillicrap, T.~P.; Beaudoin, P.; Bengio, Y.; Bogacz, R.;
  Christensen, A.; Clopath, C.; Costa, R.~P.; de~Berker, A.; Ganguli, S.;
  et~al. 2019.
\newblock A deep learning framework for neuroscience.
\newblock \emph{Nature neuroscience}, 22(11): 1761--1770.

\bibitem[{Romo et~al.(1999)Romo, Brody, Hern{\'a}ndez, and
  Lemus}]{romo1999neuronal}
Romo, R.; Brody, C.~D.; Hern{\'a}ndez, A.; and Lemus, L. 1999.
\newblock Neuronal correlates of parametric working memory in the prefrontal
  cortex.
\newblock \emph{Nature}, 399(6735): 470--473.

\bibitem[{Scellier and Bengio(2017)}]{scellier2017equilibrium}
Scellier, B.; and Bengio, Y. 2017.
\newblock Equilibrium propagation: Bridging the gap between energy-based models
  and backpropagation.
\newblock \emph{Frontiers in computational neuroscience}, 11: 24.

\bibitem[{Schuessler et~al.(2020)Schuessler, Mastrogiuseppe, Dubreuil, Ostojic,
  and Barak}]{schuessler2020interplay}
Schuessler, F.; Mastrogiuseppe, F.; Dubreuil, A.; Ostojic, S.; and Barak, O.
  2020.
\newblock The interplay between randomness and structure during learning in
  RNNs.
\newblock \emph{Advances in neural information processing systems}, 33:
  13352--13362.

\bibitem[{Schuessler et~al.(2023)Schuessler, Mastrogiuseppe, Ostojic, and
  Barak}]{schuessler2023aligned}
Schuessler, F.; Mastrogiuseppe, F.; Ostojic, S.; and Barak, O. 2023.
\newblock Aligned and oblique dynamics in recurrent neural networks.
\newblock \emph{arXiv preprint arXiv:2307.07654}.

\bibitem[{Song, Yang, and Wang(2016)}]{song2016training}
Song, H.~F.; Yang, G.~R.; and Wang, X.-J. 2016.
\newblock Training excitatory-inhibitory recurrent neural networks for
  cognitive tasks: a simple and flexible framework.
\newblock \emph{PLoS computational biology}, 12(2): e1004792.

\bibitem[{Sutton(2018)}]{sutton2018reinforcement}
Sutton, R.~S. 2018.
\newblock Reinforcement learning: an introduction.
\newblock \emph{A Bradford Book}.

\bibitem[{Vogt et~al.(2022)Vogt, Puelma~Touzel, Shlizerman, and
  Lajoie}]{vogt2022lyapunov}
Vogt, R.; Puelma~Touzel, M.; Shlizerman, E.; and Lajoie, G. 2022.
\newblock On lyapunov exponents for rnns: Understanding information propagation
  using dynamical systems tools.
\newblock \emph{Frontiers in Applied Mathematics and Statistics}, 8: 818799.

\bibitem[{Vyas et~al.(2020)Vyas, Golub, Sussillo, and
  Shenoy}]{vyas2020computation}
Vyas, S.; Golub, M.~D.; Sussillo, D.; and Shenoy, K.~V. 2020.
\newblock Computation through neural population dynamics.
\newblock \emph{Annual Review of Neuroscience}, 43: 249--275.

\bibitem[{Woodworth et~al.(2020)Woodworth, Gunasekar, Lee, Moroshko, Savarese,
  Golan, Soudry, and Srebro}]{woodworth2020kernel}
Woodworth, B.; Gunasekar, S.; Lee, J.~D.; Moroshko, E.; Savarese, P.; Golan,
  I.; Soudry, D.; and Srebro, N. 2020.
\newblock Kernel and rich regimes in overparametrized models.
\newblock In \emph{Conference on Learning Theory}, 3635--3673. PMLR.

\bibitem[{Yang and Wang(2020)}]{yang2020artificial}
Yang, G.~R.; and Wang, X.-J. 2020.
\newblock Artificial neural networks for neuroscientists: a primer.
\newblock \emph{Neuron}, 107(6): 1048--1070.

\end{thebibliography}


\newpage

\appendix

\section{Simulation details and additional simulations} \label{scn:Appendix}

\begin{figure*}[h!]
    \centering
    \includegraphics[width=0.8\textwidth]{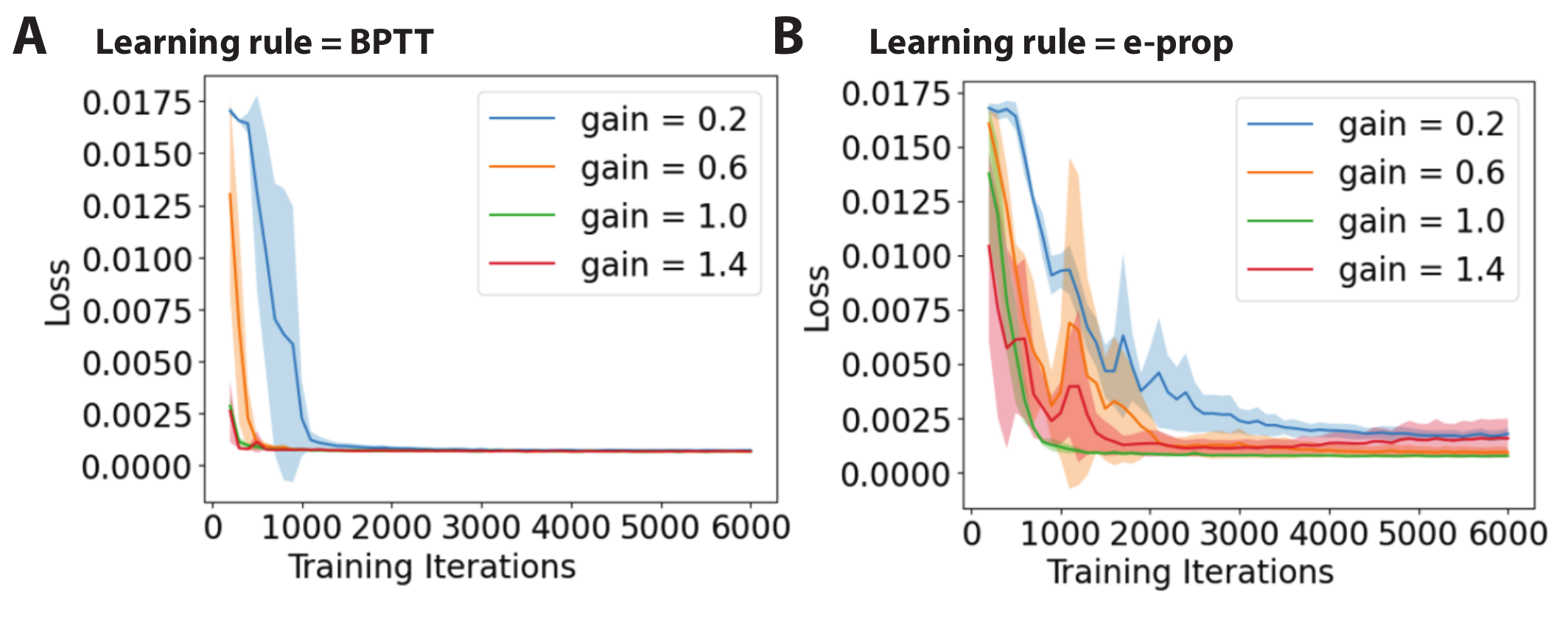}
    \caption{ Figure~\ref{fig:main_learning_curve} repeated for more intermediate gain values for A) BPTT and B) e-prop. Noticeable gap in the learning curve is observed between $gain=0.2$ with the others before convergence, even with hyperparameter tuning. Plotting convention follows that of Figure~\ref{fig:main_learning_curve}. 
    }
    \label{fig:more_gains}
\end{figure*}

Our RNN training was conducted using PyTorch using the Adam optimizer and built on the code in \cite{yang2020artificial} (see the notebook $RNN+DynamicalSystemAnalysis.ipynb$). For the 2AF and DMS tasks, we used the default Neurogym settings, while for the Romo task, we followed the implementation from~\cite{schuessler2020interplay}. E-prop was implemented in PyTorch using $hidden.detach()$, where $hidden$ is the hidden state tensor, to prevent gradient propagation across the hidden states, thereby effectively truncating the nonlocal gradient terms; this was also applied when pretraining via gradient flossing, ensuring the weight update uses location information only. Our performance evaluation utilized the learning curve, which tracks the reduction in the loss over training iterations. To give each initialization scheme a fair chance at success, we used the optimal learning rate for each initialization scheme selected from a grid of $[1e-4, 3e-4, 1e-3, 3e-3]$. By default, we used $64$ hidden neurons and a batch size of $32$, but similar trends were observed when doubling these. Each training iteration was replicated over five independent runs. All simulations were executed using Google Colab (the free version) with each run taking under 5 minutes to complete. We currently focus on recurrent weight initialization, employing standard random initialization for both the input and readout weights (initialized as in~\cite{yang2020artificial}). 

\section{Details on gradient flossing and biologically plausible learning rules} \label{scn:concepts}

Gradient flossing, originally proposed in~\cite{engelken2024gradient}, addresses the problem of exploding and vanishing gradients in recurrent neural networks by regularizing Lyapunov exponents. This method has several variants, including applying gradient flossing intermittently during training or as a pretraining step. In this work, we adopt the latter approach, where the network is pretrained with the flossing loss to push the Lyapunov exponents $\lambda_i$ toward zero:  
\[ L_{flossing} = \sum_i^k \lambda_i^2. \] This stabilization of Lyapunov exponents ensures both forward and gradient dynamics remain well-behaved. Additionally, as mentioned earlier, locality constraints were enforced during the pretraining phase. 

We also explain the approximation mechanisms used by each biologically plausible learning rule. For a detailed explanation, readers are encouraged to consult the referenced works. We start by expressing the gradient using the real-time recurrent learning (RTRL) factorization, which is a causal equivalent to the backpropagation through time (BPTT) gradient factorization:

\begin{eqnarray} \frac{\partial L}{\partial W_{h, ij}} &=&\sum_{l,t}\frac{\partial L}{\partial h_{l,t}} \frac{\partial h_{l,t}}{\partial W_{h, ij}}, \label{eqn:sum_tl} \end{eqnarray}

The key challenge with RTRL, in terms of both biological plausibility and computational feasibility, lies in the term $\frac{\partial h_{l,t}}{\partial W_{h, ij}}$, which tracks the recursive dependencies of $h_{l,t}$ on $W_{h,ij}$ through the network’s recurrent connections. This term is calculated recursively as follows:

\begin{align}
\frac{\partial h_{l,t}}{\partial W_{h, ij}} &= \frac{\partial h_{j,t}}{\partial W_{h,ij}} + \sum_m \frac{\partial h_{l,t}}{\partial h_{m,t-1}} \frac{\partial h_{m,t-1}}{\partial W_{h,ij}} \cr 
&= \frac{\partial h_{l,t}}{\partial W_{h,ij}} + \frac{\partial h_{l,t}}{\partial h_{l,t-1}} \frac{\partial h_{l,t-1}}{\partial W_{h,ij}} \cr 
& \quad + \underbrace{\textstyle \sum_{m\neq l} W_{h,lm} f'(h_{m,t-1}) \frac{\partial h_{m,t-1}}{\partial W_{h,ij}}}_{\text{\normalsize involving all weights $W_{h,lm}$}}. \label{eqn:s_triple}
\end{align}

This dependency introduces a significant challenge for biological plausibility since $\frac{\partial h_{l,t}}{\partial W_{h, ij}}$ includes nonlocal terms. Specifically, updating each weight $W_{h,ij}$ would require knowledge of all other weights in the network, which is biologically unrealistic. \textbf{For a learning rule to be biologically plausible, all the information required to update a synaptic weight must be locally accessible at the synapse. However, how neural circuits could make such global information on the weights and activity of the entire network available to individual synapses remains an open question.}

To address this, learning rules like \textbf{e-prop}~\cite{bellec2020solution} and its equivalent, \textbf{RFLO}~\cite{murray2019local}, approximate the gradient by truncating these nonlocal terms in Eq.~\ref{eqn:s_triple}. This ensures that weight updates follow a biologically plausible three-factor framework, where updates depend only on local pre- and post-synaptic activity along with a top-down instructive signal (e.g., neuromodulators):

\begin{align}
\widehat{\frac{\partial h_{l,t}}{\partial W_{h, ij}}} &=
\begin{cases}
\frac{\partial h_{i,t}}{\partial W_{h,ij}} + \frac{\partial h_{i,t}}{\partial h_{i,t-1}} \widehat{\frac{\partial h_{i,t-1}}{\partial W_{h,ij}}}, & \text{if } l = i \\
0, & \text{if } l \neq i
\end{cases}
\end{align}

This approximation greatly simplifies the computation compared to the full tensor in Eq.~\ref{eqn:s_triple} and preserves the locality constraints so that synaptic updates use only signals locally available to that synapse. As mentioned, this truncation can be implemented in PyTorch using $h.detach()$, which prevents gradients from propagating through the recurrent weights.

\end{document}